\documentclass{article}
\usepackage[utf8]{inputenc}

\usepackage{color}
\usepackage{hyperref}
\usepackage{amsmath}
\usepackage[nameinlink,capitalise]{cleveref}

\usepackage{enumitem}
\usepackage{graphicx}
\usepackage{subcaption}
\graphicspath{{./images/}}
\usepackage{amsmath,amssymb,amsxtra,amsfonts}
\usepackage{comment}
\usepackage{bm}
\usepackage[linesnumbered,ruled]{algorithm2e}
\usepackage{csquotes}
\usepackage{biblatex}
\usepackage[utf8]{inputenc}
\usepackage{dirtytalk}
\usepackage{multirow}
\usepackage[a4paper, margin=0.95in]{geometry}

\newcommand{\vecxi}{\bm{x}_{i}}

\def\bse{\begin{eqnarray*}}
\def\ese{\end{eqnarray*}}
\def\bae{\begin{eqnarray}}
\def\eae{\end{eqnarray}}
\def\bsq{\begin{equation*}}
\def\esq{\end{equation*}}
\def\bq{\begin{equation}}
\def\eq{\end{equation}}

\def\argmin{\mbox{argmin}}

\def\0{{\bf 0}}
\def\1{{\bf 1}}

\def\bq{\begin{equation}}
\def\eq{\end{equation}}

\def\log{\hbox{log}}

\def\squarebox#1{\hbox to #1{\hfill\vbox to #1{\vfill}}}

\def\bse{\begin{eqnarray*}}
\def\ese{\end{eqnarray*}}
\def\bae{\begin{eqnarray}}
\def\eae{\end{eqnarray}}
\def\bsq{\begin{equation*}}
\def\esq{\end{equation*}}
\def\bq{\begin{equation}}
\def\eq{\end{equation}}

\def\boxit#1{\vbox{\hrule\hbox{\vrule\kern6pt\vbox{\kern6pt#1\kern6pt}\kern6pt\vrule}\hrule}}

\begin{document}

\title{\textbf{Lockout: Sparse Regularization of Neural Networks}}
\author{
	\textbf{Gilmer Valdes$^*$} \\
	Department of Radiation Oncology \\
	Department of Epidemiology and Biostatistics \\
        University of California San Francisco, CA 94143, USA \\
         \textit{gilmer.valdes@ucsf.edu}
  \and
	\textbf{Wilmer Arbelo$^*$} \\
        Department of Radiation Oncology\\
        University of California San Francisco, CA 94143, USA \\
        \textit{wilmer.arbelogonzalez@ucsf.edu}
  \and
	\textbf{Yannet Interian}  \\
	M.S. in Data Science Program \\
        University of San Francisco, San Francisco, CA 94105, USA \\
        \textit{yinterian@usfca.edu}
   \and
	\textbf{Jerome H. Friedman}  \\
        Department of Statistics \\ 
        Stanford University, Stanford, CA 94305, USA \\
        \textit{jhf@stanford.edu}  \\
   	\\
        \textbf{$^*$}\small{equal authors contribution} \\
}
\date{May 12, 2021}

\maketitle

\begin{abstract}
Many regression and classification procedures fit a parameterized function $f(\bm{x};\bm{w})$ of predictor variables
$\bm{x}$ to data $\{\bm{x}_{i},y_{i}\}_1^N$ based on some loss criterion
$L(y,f)$. Often, regularization  is applied to improve accuracy by placing a constraint $P(\bm{w})\leq t$ on the values
of the parameters $\bm{w}$. Although efficient methods exist for finding solutions to these constrained optimization problems for all values of $t\geq0$ in the special case when $f$ is a linear function, none are available when $f$ is non-linear (e.g. Neural Networks). Here we present a fast algorithm that provides all such solutions for any differentiable function $f$ and
loss $L$, and any constraint $P$ that is an increasing monotone
function of the absolute value of each parameter. Applications involving
sparsity inducing regularization of arbitrary Neural Networks are discussed.
Empirical results indicate that these sparse solutions are usually superior to their dense counterparts in both accuracy and interpretability. This improvement in accuracy can often make Neural Networks competitive with, and sometimes superior to, state-of-the-art methods in the analysis of tabular data.
\end{abstract}

\section{Introduction}

Neural Networks (NNs) have become popular in recent years  for problems involving image classification and natural language processing \cite{krizhevsky2012imagenet,simonyan2014very,seide2011conversational}. In the analysis of tabular data, however, they still lag behind other competitors \cite{friedman2001greedy, arik2019tabnet}. NNs usually have many hidden units and layers that make them very expressive models but prone to over-fitting in problems that involve substantial amount of noise, many variables or limited data \cite{srivastava2014dropout}. Many important problems, especially those in medicine, are of this latter type. For instance, genome wide studies usually involve thousands of genes but the number of patients with outcomes is in the hundreds.  Thus, they are not amenable to the advantages of representational learning and feature design that NNs provide. 

For these problems, methods like L2, L1, Elastic Net or the Group Lasso regularization are quite popular \cite{hastie2009elements, zou2005regularization, tibshirani1996regression, meier2008group}. They involve a constrained optimization problem of the type
\begin{equation}
\displaystyle \hat{\bm{w}} = \argmin_{\bm{w}}  \mathbb{E}_{\bm{x},y} \left( L\left[y, f\left(\bm{x};\bm{w}) \right)\right]\right) \qquad\textrm{s.t.}\quad  P(\bm{w})\leq t_0.
\label{eqn1}
\end{equation}
Here $L$ is a loss function, $y$ is the outcome, $\bm{x}$ the covariates, $\bm{w}$ the parameters, $f$ is a linear function, $t_0 \geq 0$  and  $P(\bm{w})$ is a function that monotonically increases with the absolute value of each of the parameters.  These methods are able to improve performance by providing regularization that reduces the magnitude of the coefficients, thereby controlling the variance of their estimates. Additionally, some (e.g. L1 or Group Lasso) impose sparsity or feature selection by setting many coefficient values to zero. This is of paramount importance because often interest is not only in performance but also in understanding the factors driving it. For instance, in predicting cancer survival using microarray DNA expression data, understanding the genes and pathways responsible for the outcome can lead to the design of drugs to target them. Methods like Lasso or Group Lasso, however, can only be applied to models that have a linear dependence on the parameters and can perform poorly when this is not the case.

For these circumstances one might attempt to apply Neural Networks (NNs) in combination with L1, L2 or Group Lasso regularization.  Unfortunately, available methods require specifying a value for the regularization hyperparameter $t_0$ (per layer, node or the whole network). As good values for these regularization parameters are seldom known, they must be estimated by repeatedly evaluating trial solutions through  cross-validation. Given the computational cost of optimizing NNs, cross-validating a wide range of values for these hyperparameters is generally not feasible.

Even for linear models with some constraint functions, solving the optimization problem in Eq.~(\ref{eqn1}) for different values of these hyperparameters can be too slow to be practical. This issue was addressed by introducing path seeking algorithms \cite{best1996algorithm,osborne2000lasso,efron2004least, friedman2010regularization, tibshirani2011solution}. We can see in Eq.~(\ref{eqn1}) that given $L, f,$ and $P$, $\hat{\bm{w}}$ is a function only of $t_0$. If the solution to the optimization problem at $t_0$, $\hat{\bm{w}}^{t_0}$, is known, we can proceed to find the solution at $t = t_0 + \Delta t$, $\hat{\bm{w}}^{t_0 + \Delta t}$, by taking small steps from $\hat{\bm{w}}^{t_0}$ that satisfy the new constraint, $t_0 + \Delta t$ . Noticing that $\hat{\bm{w}}^{t_0} = \mathbf{0}$ when $t_0 = 0$, to find the optimal value of $\hat{\bm{w}}$ we can repeatedly increase $t_0$ by small increments and stop when the validation error minimizes. The collection of all the tuples $\{\hat{\bm{w}}^t, t\}$ become the path.

Two issues prevent us from using the algorithms cited above for NNs:  1) they require the relationship between parameters and output to be linear and 2) they generally start at $t_0=0$, $ \bm{w}^{0} = \mathbf{0}$ which is not possible for NNs because this is a bad local minimum. Here, we extend ideas previously developed by Friedman \cite{friedman2012fast} to create a fast path seeking algorithm (\say{Lockout}) that addresses these limitations. Lockout can easily be applied to the optimization of any Neural Network  architecture (or other arbitrary nonlinear function) based on any differentiable loss criterion and any convex or non-convex parameter constraint that is an increasing monotone function of the absolute value of each parameter. These include the L1, L2, Group Lasso, Elastic Net convex constraints, as well as non-convex ones such as SCAD \cite{fan1999variable}, MC+ \cite{zhang2010nearly}, etc.). When applied to Neural Networks, Lockout provides a regularization path for any network architecture and for parameters in any layer. As such, it can be used not only to provide feature selection, but also architecture selection. In terms of computational cost, it only adds $o(\bm{M}\log\bm{M})$ extra steps compared to the standard back propagation algorithm, where $\bm{M}$ is the number of parameters included in the regularization. It can be implemented by simply applying minor modifications, involving only a few lines of code, to any existing optimization algorithm (e.g. Adam \cite{kingma2014adam},  Stochastic Gradient Descent, etc.).

\section{Related Work}

There is a body of work aimed at regularizing NNs by applying penalties or constraints to the values of the parameters. \cite{feng2017sparse, scardapane2017group, alvarez2016learning, bach2017breaking}. These approaches apply regularization to achieve feature selection in the first layer or architecture selection in subsequent layers \cite{feng2017sparse, scardapane2017group, alvarez2016learning, bach2017breaking}. However, they require the specification of a value for the hyper-parameter controlling the strength of the penalty or constraint. Since a good value is seldom known in advance, it must be obtained by repeatedly applying cross validation. Additionally, they incorporate the constraint as a Lagrangian in the main loss function \cite{feng2017sparse, scardapane2017group, alvarez2016learning, bach2017breaking}. This technique, although widely used, prevents effective training of NNs in a subtle way, as we will explain below.

If strong regularization is imposed and the Lagrangian term is incorporated in the loss function, the gradient from the regularization eventually becomes dominant, and the procedure stops learning \cite{Park2018AchievingSR}. This is especially accentuated in NNs because (except for the first layer) the gradients shift during training together with the residuals. This behaviour makes it harder to find an optimal hyperparameter that is appropriate for all the training phases \cite{Park2018AchievingSR}, even when batch normalization is applied \cite{Park2018AchievingSR}. This may be a reason why regularization methods like L1, L2 or Group Lasso are still not popular in the NN community, where early stopping and dropout dominate \cite{srivastava2014dropout,wager2013dropout}.  This problem has prompted ad hoc approaches that vary the strength of the regularization constraint at different stages of the training cycle \cite{Park2018AchievingSR}. Lockout avoids this problem as it is not based on the lagrangian formulation. 

Recently, Lemhadri et al. \cite{lemhadri2021lassonet} introduced an algorithm (LassoNet) that applies feature selection to NNs by 1) creating a linear residual connection between the input layer and the output, 2) imposing L1 regularization to the parameters in the residual linear connection and 3) adding as an extract constraint that the absolute value of the parameters from the non-linear portion of the NN be smaller than the absolute value of the parameters from the corresponding linear component multiplied by a positive constant. The combination of these three steps results in an elegant mathematical solution to the problem, but has several inherent limitations.

First, it is restricted to a special type of (residual) Neural Network that requires a linear connection to the output. Many important architectures today are not residual. Second, the method used to induce sparsity (combination of steps 2 and 3 above) can only be applied to the first layer of the network and not to subsequent layers. Also, the method relies on coupling the strength of the regularization for the parameters associated with the linear connection to those in the non-linear (residual) portion of the network. As such, it will be inhibited from accurately approximating functions with variables that have a strong non-linear component in the absence of a correspondingly strong linear counterpart. This effect is illustrated below using datasets with these characteristics, for which Lockout achieves significantly better performance.

\section{Notation}

Let $\vecxi \in \mathcal{X}$ be  a $p$ dimensional
feature vector, and $y_i \in \mathcal{Y}$ be its corresponding outcome for the observation $i$. We will indicate vectors and matrices with bold font letters ({e.g. $\bm{a}, \bm{b}, \dots$}) and their components with italic ($a_j, b_j, \dots$). $NN$ will refer to a generic Neural Network and $\bm{w}$ the set of all its parameters. The notation $\bm{w}_{j}$ refers to a generic parameter $j$ in some node and layer. We do not index the specific node and layer because it will complicate the notation without any direct benefit. If no index is provided for a lower case letter (e.g. $w$), we are referring to a generic parameter. 

In this paper, we define the regularization path to be the sequence $\{ \hat{\bm{w}}^t\}$ for $t \in [0, t^*]$, where $\hat{\bm{w}}^t$ is a solution to the optimization problem
\begin{equation}
\displaystyle \hat{\bm{w}}^t = \argmin_{\bm{w}}    \mathbb{E}_{\bm{x},y} \left(L\left(y, f(\bm{x};\bm{w}) \right) \right) \qquad\textrm{s.t.}\quad P(\bm{w})\leq t
\label{eqn2}
\end{equation}
and $t^*$ is the value of $P(\hat{\bm{w}})$ corresponding to an unconstrained solution to Eq.~\ref{eqn2}. That is, $t^* = P(\hat{\bm{w}}^{+\infty})$.  

\section{Lockout Intuition}
\label{sec:LO-Intuition}

The justification for Lockout is straight forward. At every iteration we 1) linearize the loss function and constraint, 2) solve the corresponding linear programming problem and 3) apply these steps iteratively. This linearization technique is a common convex optimization procedure called Successive Linear Programming (SLP) or  Method of Approximation Programming (MAP) and its origins can be traced to the Frank Wolf method \cite{frank1956algorithm}. It has been studied by various authors and the interested reader is referred to \cite{palacios1982nonlinear, meyer1970validity, meyer1968solution, luo1999extensions} for theoretical guarantees and convergence theorems. 

The operation of this procedure is also intuitive. We start at an unconstrained solution of Eq. (\ref{eqn2}), $\hat{\bm{w}}^{t^*}$, that by definition is on the regularization path. We then walk it backwards by repeatedly decreasing the constraint $t^*$ in small increments. At each such step, using the updates provided by an optimizer, we apply the following in order:

\begin{enumerate}
\item Update parameters for which penalization is $0$ as these do not affect the constraint. 
\item Update parameters for which decreasing their absolute value (and as such their penalty) decreases the loss.
\item Update the most important parameters (as defined by Lockout) in the direction provided by the optimizer, which will increase their absolute value but decrease the loss function.
\item Update the rest of the parameters contrary to the direction provided by the optimizer, which will decrease their absolute value but increase the loss. 
\end{enumerate}

The key ingredient is that the increase in the penalty function provided by step 3 needs to be balanced by steps 2 and 4. Equally, by letting the most important parameters change in the direction that decreases the loss function (step 2 and 3), they will compensate the increase in the same when the parameters in step 4 (less important parameters) are moved in the contrary direction. A general formulation for Lockout is provided in the Appendix, together with the pseudocode for the case when the penalization is provided at the first layer and the optimization algorithm is stochastic gradient descent. Note that Lockout does not solve the minimization problem using the Lagrangian formulation.  It rather requires the weight parameters to compete for which ones are allowed to increase in absolute value. This is a unique characteristic of our method that insures its effectiveness when applied to NNs.

\section{Model Selection and Initialization}

In the previous section, we have described Lockout as traversing the entire regularization path. Starting at the unconstrained solution, the strength of the constraint is successively increased until all parameter values are zero. The path point $\hat{\bm{w}}^t$ with minimum estimated prediction error (or some other criterion) then defines the selected model. Computationally, this strategy requires solving for the unconstrained solution and then passing over many dense undesirable ones along the path in the search for an appropriate sparse solution. Although this strategy can be thorough, a more efficient alternative is to start Lockout at the usual cross-validated early stopping solution. The idea is that it is unlikely that the solution with the best performance would need to be less constrained than the early stopping solution. Naturally, this solution would not be on the regularization path, so slight changes to Lockout implementation are required.
In particular, we need to:
\begin{itemize}
    \item Compute the value of the constraint $t=t_0$ corresponding to the early stopping solution.
    \item Train the NN using Lockout with a constant constraint value $t=t_0$ until it reaches the regularization path.
    \item Starting from the path point found in the previous step, traverse the path in the normal way by iteratively reducing $t$.
\end{itemize}

This new strategy opens the door for the use of Lockout with any existing NN. In any particular application, a researcher can obtain a NN solution using their method of choice and then apply Lockout to that solution in an attempt to find one with improved accuracy and/or sparsity. 

Regardless of the initial model, an important consideration when using Lockout, especially for classification tasks, is the accuracy-sparsity trade-off in model selection. Usually, the selected solution is taken to be the one that minimizes the estimated validation prediction risk. It is often the case with Lockout that there are many solutions with prediction risk estimates only very slightly larger than the minimizing one, but that are much sparser. These sparser solutions might be preferred for interpretability or even accuracy if strong sparsity is a prior belief.
\begin{figure}[t]
    \centering
    \includegraphics[width=0.7\textwidth]{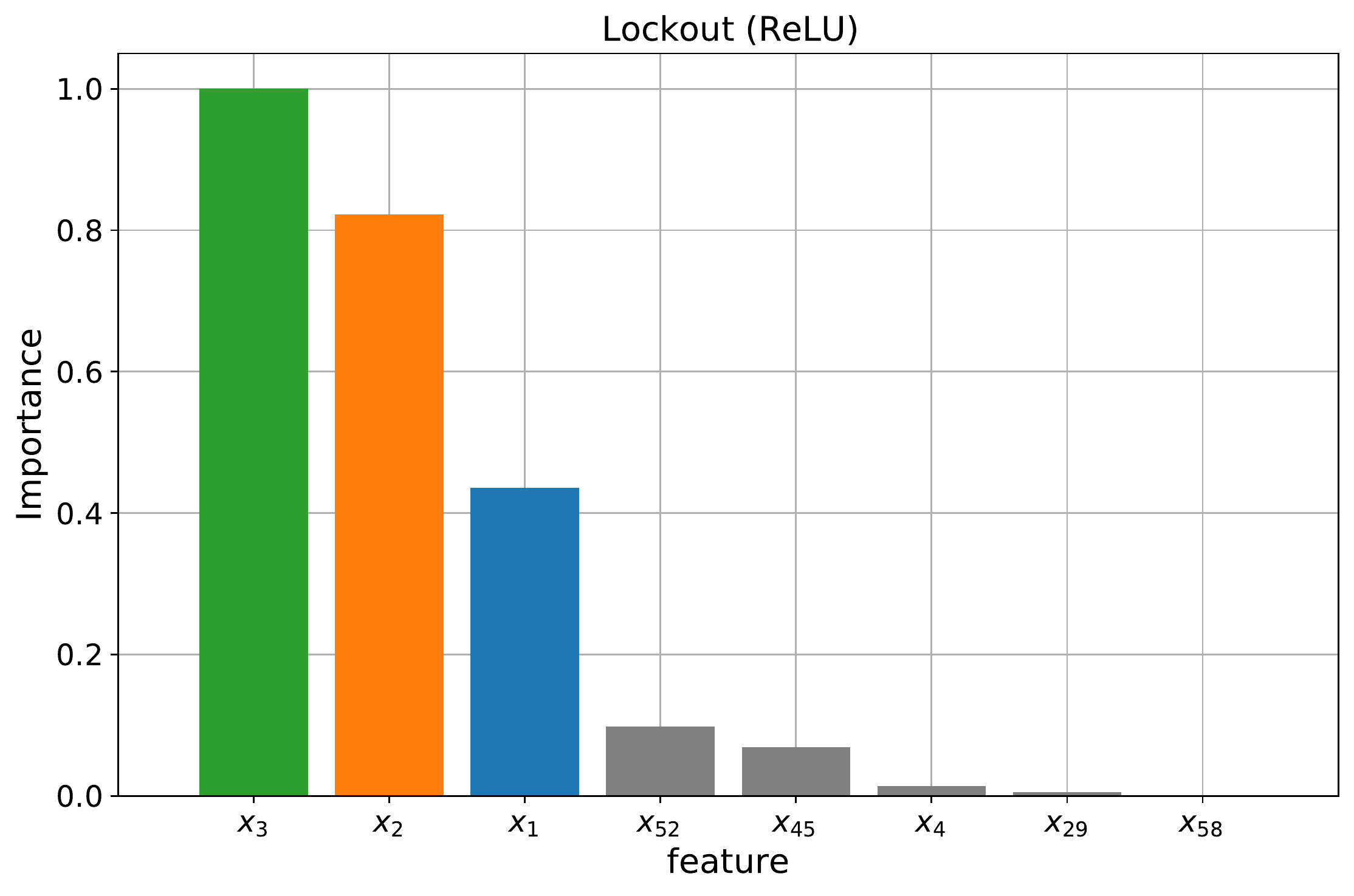}
     \caption{Feature importance, computed as $|w_i|$, for the model found by Lockout at the validation minimum. True values of $\{a_i\}_1^3$ are $a_1=0.5$, $a_2=0.75$, and $a_3=1.0$}.
     \label{fig:feat_import_LO_b}
\end{figure}

\begin{figure}[t]
     \centering
     \includegraphics[width=\textwidth]{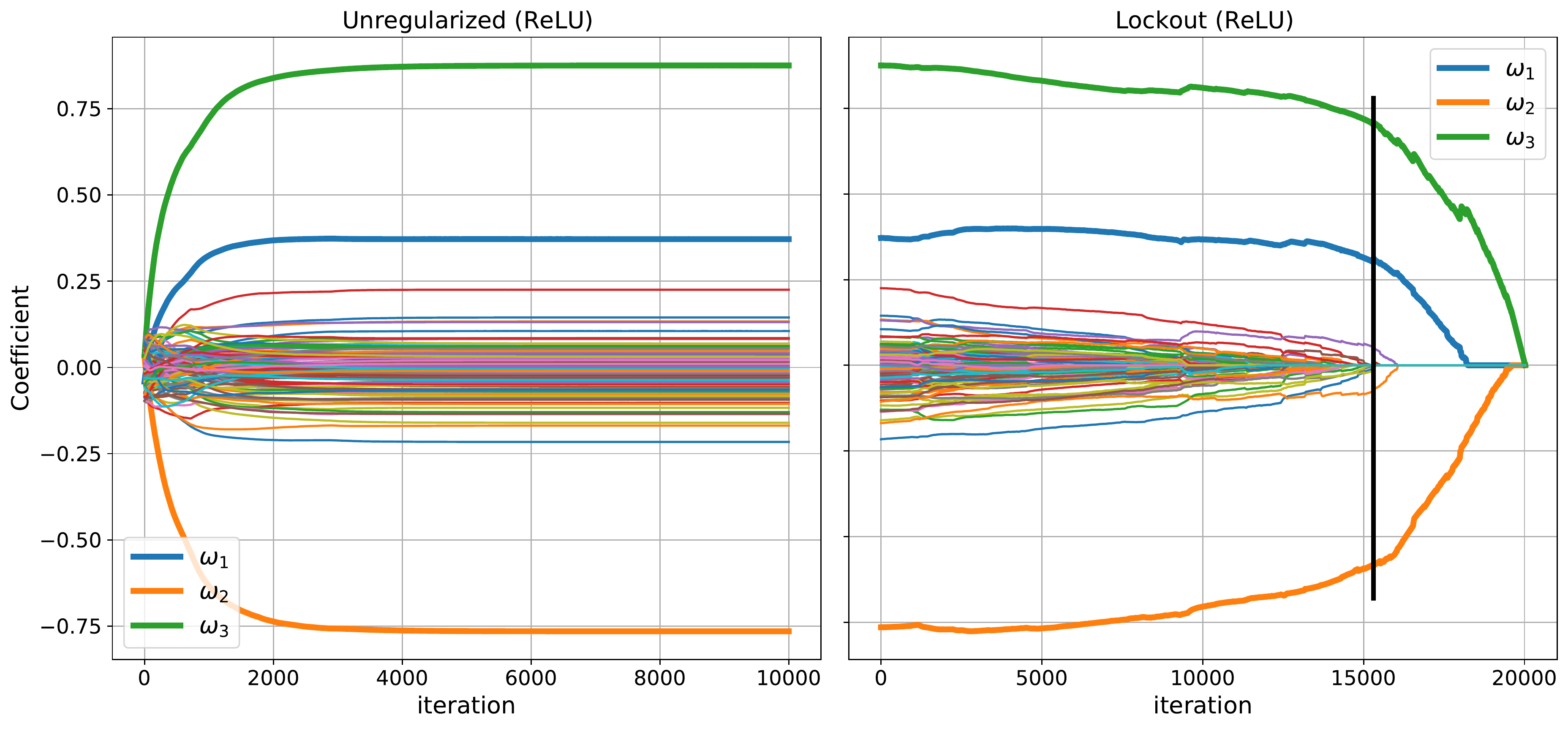}
     \caption{First-layer coefficients vs. number of iterations when the NN is trained without regularization (left) and with Lockout (right). For the latter, the vertical line represents the position of the validation minimum.}
     \label{fig:w_vs_iters_b}
\end{figure}

\section{Experiments}

Here, we have applied Lockout to both synthetic and real datasets in  classification and regression tasks. We had four main empirical objectives: 
\begin{itemize}
  \item show that Lockout can select the correct features in non-linear problems.
  \item show how Lockout can improve accuracy of NNs.
  \item establish those problems for which Lockout might be preferred over existing methods.
  \item evaluate whether Lockout can make NNs competitive in problems for which they are usually considered inferior.
\end{itemize}

\subsection{Synthetic Data I}
\label{sec:SynDataI}

In this section, we apply Lockout with L1 regularization to four different synthetic data sets $\{\bm{x}_i, y_i\}_1^N$. The objective is to show that Lockout can provide feature selection while improving the accuracy of the NN. 
The data sets for each experiment contained 500 observations with 100 features. They were each partitioned into separate training, validation, and testing subsets. 
Input variables $\{\bm{x}_i\}^{500}_1$ were randomly drawn from a uniform distribution $\bm{x}_i\sim U[0,1)$. 
Targets $\{y_i\}^{500}_1$ were generated as $y_i=g(z_i)$, where $z_i = a_1x_{i1} - a_2x_{i2} + a_3x_{i3}$ involves only 3 out of the 100 features and constants $\{a_i\}_1^3$ were respectively set to 0.5, 0.75 and 1.0. 
A different activation function $g(z_i)$ was used for each data set. 
In particular, $g(z_i) = z_i$, $g(z_i) = \mathrm{ReLU}(z_i)$, $g(z_i) = \tanh{(z_i)}$, and 
$g(z_i) = \mathrm{Sigmoid}{(z_i)}$ were used to generate the four datasets.
Gaussian noise was added to both the training and validation subsets, with a signal-to-noise ratio of 1. No noise was added to the testing subset. 

In the first experiment, we use one hidden layer with a single node architecture, followed by the same activation $g(z_i)$ used to generate the corresponding targets. More complex architectures and realistic scenarios are used in other experiments.
The NN is first trained, and both the unconstrained (training data minimum) and the early stopping (validation minimum) solutions are obtained.
The unconstrained solution was assumed to be reached if changes in the training loss function were smaller than $10^{-5}$ for 20 consecutive iterations.
This solution is then used as the starting point to train the NN with Lockout.
The learning rates used in the unregularized and Lockout training were $5\times10^{-3}$ and $10^{-2}$, respectively.
Stochastic gradient descent (SGD) was used as the optimization algorithm in both experiments.

\begin{figure}[ht]
     \centering
     \includegraphics[width=\textwidth]{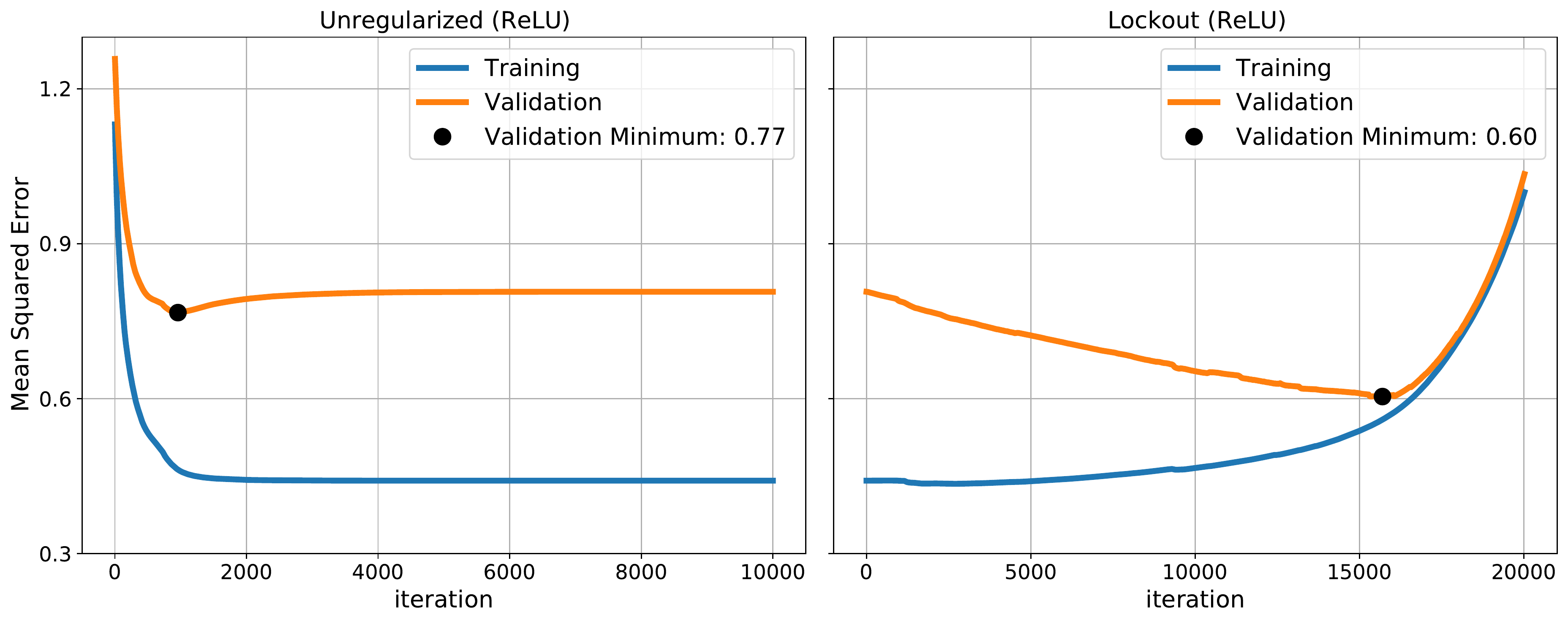}
     \caption{Training and validation mean squared errors vs. number of iterations when the NN is trained without regularization (left) and with Lockout (right). Values of the validation minimums found in each case is also shown in the figure.}
    \label{fig:loss_vs_iter_b}
\end{figure}

Figures \ref{fig:feat_import_LO_b} and \ref{fig:w_vs_iters_b} show that Lockout provides strong feature selection. For instance, at the validation minimum, only 7 out of the 100 features are non-zero. Among the seven non-zero features, the three that generated the output have the biggest importance. A similar result was found for the other examples. Lockout also provided significantly improved accuracy over the usual early stopping solutions for all activation functions used, as can be seen in Figure \ref{fig:loss_vs_iter_b} and Table \ref{tab:datasets1_4}.

\setlength{\arrayrulewidth}{0.2mm}
\renewcommand{\arraystretch}{1.5}
\begin{table}
    \centering
    \begin{tabular}{c|c|c|c|c|}
        \cline{2-5}
          & \multicolumn{2}{|c|}{Validation Error} & \multicolumn{2}{|c|}{Testing$^*$ Error} \\
        \cline{2-5}
                  & Early Stopping & Lockout(L1) & Early Stopping & Lockout(L1) \\
        \hline
        \multicolumn{1}{|c|}{Linear}  & 0.783 & 0.727 & 0.460 & 0.204 \\
        \multicolumn{1}{|c|}{ReLU}    & 0.868 & 0.770 & 0.652 & 0.414 \\
        \multicolumn{1}{|c|}{Tanh}    & 0.874 & 0.827 & 0.665 & 0.550 \\
        \multicolumn{1}{|c|}{Sigmoid} & 0.855 & 0.826 & 0.643 & 0.566 \\
        \hline
    \end{tabular}
    \caption{Relative root mean squared error, defined as $\sqrt{1-R^2}$, for the validation and testing subsets corresponding to the early stopping and Lockout models. In all cases, the NN used had only one node with the same activation function that generated the data. Lockout considerably improves the early stopping solution. $^*$Note that testing subsets were generated with no noise added. }
    \label{tab:datasets1_4}
\end{table}

\subsection{Synthetic Data II}

In this section, we apply L1 based Lockout to two different nonlinear synthetic data sets. The goal is to illustrate its performance in the presence of strong nonlinear dependence of $y$ on $\bm{x}$, with or without a corresponding linear dependence, and compare it to Lasso \cite{tibshirani1996regression} or the more recent LassoNet \cite{lemhadri2021lassonet}. 

For each of the datasets, 1000 observations and 200 features were generated for training, validation, and testing. 
Input variables $\{\bm{x}_i\}^{1000}_1$ were drawn from a uniform distribution $\bm{x}_i\sim U[0,1)$ .
The target values $\{y_i\}^{1000}_1$ for the first experiment were generated using the well researched Friedman function \cite{friedman1991multivariate}, 
\begin{equation}
    y_i = 10\sin{(\pi x_{i1}x_{i2})} + 20(x_{i3}-0.5)^2 + 10x_{i4} + 5x_{i5}.
    \label{eq:friedman}
\end{equation}
For the second experiment the last two linear terms were omitted.
Gaussian noise was added to the training and validation subsets, with a signal-to-noise ratio of 0.5. No noise was added to the testing subset.

Two NN architectures were explored in these experiments. The first one was a 2-layer fully connected architecture with 10 nodes in each hidden layer involving ReLU activation functions, and one node in the output layer. The second architecture was the same as the first, except for the addition of an extra residual linear connection between the input features and the output node (ResNet). The latter architecture was included because the method LassoNet requires this extra residual connection to provide regularization \cite{lemhadri2021lassonet}. The unconstrained solution in each case was found using the Adam optimizer with a learning rate of $10^{-3}$. The same convergence condition as in the previous experiments was used. For Lockout, SGD with a learning rate of $10^{-2}$ was employed.

\begin{figure}[ht!]
    \centering
    \includegraphics[width=\textwidth]{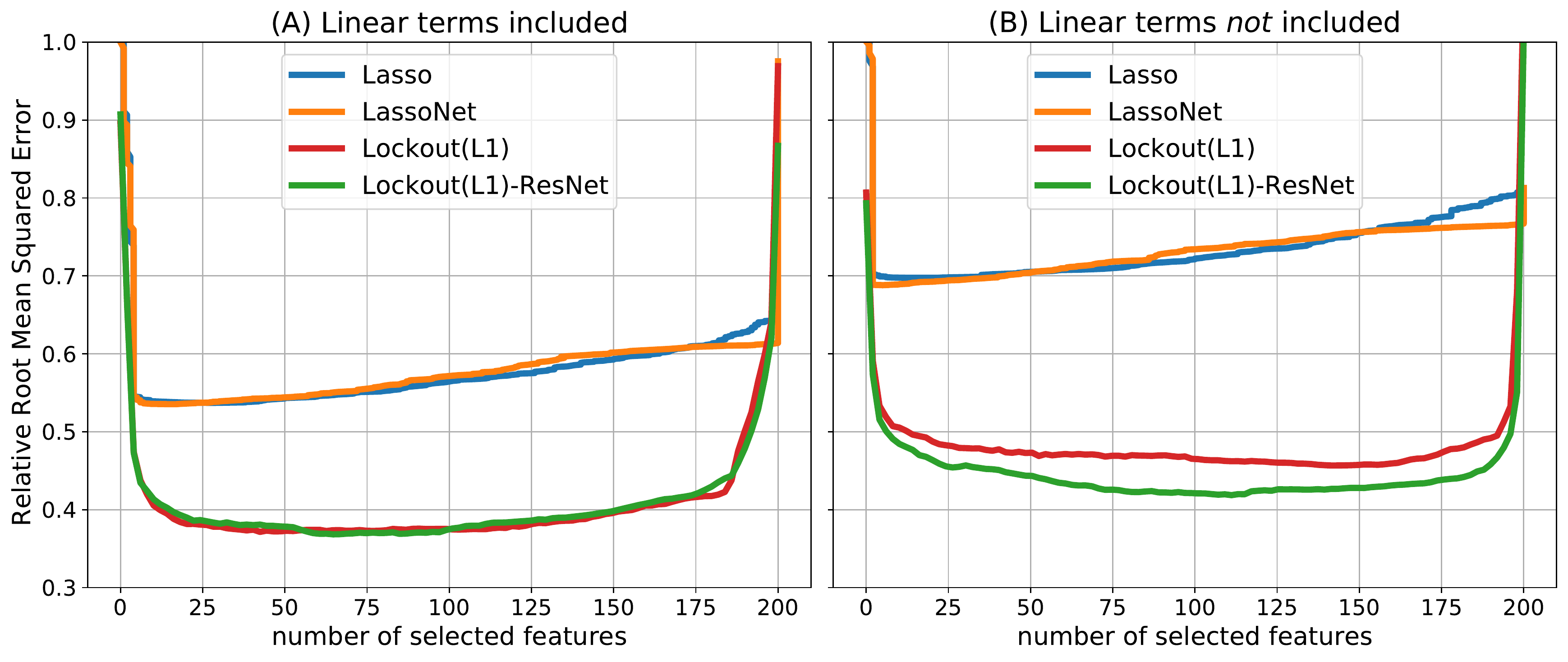}
    \caption{Relative root mean squared error vs. number of selected features for the testing subset. In addition to the two NN architectures previously described, results from Lasso \cite{tibshirani1996regression} and LassoNet \cite{lemhadri2021lassonet} are also shown. Results shown are over one run, after selecting the best NN using the validation set. 2 fully connected layers with 10 nodes in the hidden layers with or without the residual connection was used.}
    \label{fig:error_vs_density_ab}
\end{figure}

Figure~\ref{fig:error_vs_density_ab} shows relative root mean squared error vs. number of selected features for the testing subset for Lasso (linear model)~\cite{tibshirani1996regression}, LassoNet~\cite{lemhadri2021lassonet} (ResNet model), Lockout with the fully connected model and Lockout with the residual model. The results show that the strong non-linear components in the data generating function, in the absence of corresponding strong linear counterparts, cause Lockout to significantly outperform Lasso or LassoNet in these examples. Lasso is only able to model linear functions, whereas LassoNet couples the regularization of the linear part with that of the non-linear dependence. As such, they both are at a disadvantage when the variables responsible for the linear terms are not the same as those responsible for the non-linearity. By not coupling the regularization of non-linear dependencies to those of the corresponding linear ones, Lockout avoids this limitation.

\subsection{Real Datasets}

In this section, we apply L1 based Lockout to various classification and regression problems involving real-world datasets.
The goal is to show that Lockout can often improve NN's performance, making it competitive with and sometimes superior to traditional state-of-the-art methods for tabular data such as Lasso or Gradient Boosting (GB). Additionally, we present a novel application of Lockout to cancer research, where representational learning is a must. 

\subsubsection*{Experiment 1}
We analyzed three microarray DNA datasets related to different types of cancer. Historically, NNs have been substantially outperformed by other state-of-the-art methods in these settings. The problems are summarized in Table~\ref{tab:exp1}. 
Each example is a binary classification problem where the target variable $y_i=1$ if the patient has cancer and $y_i=0$ otherwise.
Given the small number of observations, datasets were partitioned into 100 different training, validation, and testing subsets randomly drawn with  $60\%$, $20\%$ and $20\%$, of the data, respectively.
Reported metrics for each method are the average and standard deviation over all the partitions. We also performed a paired T-test to compare performance of the algorithms against Lockout with L1 regularization. 

\renewcommand{\arraystretch}{1.4}
\begin{table}[ht!]
    \centering
    \begin{tabular}{|c|c|p{13.8cm}|}
        \hline
        Id & Size & \multicolumn{1}{|c|}{Description} \\
        \hline
        
        \multirow{2}{*}{1} & \multirow{2}{*}{120$\times$54675} & Collected from 32 pairs of lung tumor and adjacent normal tissue specimens from non-smoking female adenocarcinoma patients~\cite{lu2015identification}. \\
        \hline
        
        \multirow{2}{*}{2} & \multirow{2}{*}{156$\times$54352} & Obtained from 39 patients with squamous cell carcinoma of the uterine cervix, before and during fractionated radiotherapy~\cite{iwakawa2007radiation}.\\
        \hline
        
        \multirow{2}{*}{3} & \multirow{2}{*}{3744$\times$228} & Obtained from 1698 lung cancer and 207 non-cancer serum samples from National Cancer Center Biobank and 1998 non-cancer serum samples from Yokohama Minoru Clinic~\cite{asakura2020mirna}.\\
        \hline
        
    \end{tabular}
    \caption{Microarray DNA datasets used in Experiment 1. Dataset size is given as $N\times p$, where $N$ is the number of observations and $p$ the number of features. More details can be found in the corresponding references.}
    \label{tab:exp1}
\end{table}

A grid search over relevant hyperparameter values was performed for each algorithm and the validation subset used for model selection. In particular, we tuned 1) the regularization strength for Lasso, 2) the number of trees, maximum depth of a tree, and shrinkage coefficient for Gradient Boosting, and 3) the learning rate for early stopping and Lockout(L1) NNs. 
In all examples, we used a 2-layer fully connected architecture with 5 nodes in the hidden layer using ReLU activation functions. There were two nodes in the output layer. 
The traditional early stopping and Lockout(L1) solutions (at their respective validation minimums) were found using Adam and SGD, respectively, with cross-validated learning rate values.
Lasso calculations were computed with its Scikit-Learn implementation~\cite{scikit-learn} while GB's with XGBoost package~\cite{Chen:2016:XST:2939672.2939785}.

\setlength{\arrayrulewidth}{0.2mm}
\setlength{\tabcolsep}{5pt}
\renewcommand{\arraystretch}{1.4}
\begin{table}[ht!]
    \centering
    \begin{tabular}{|c|c|c|c|c|}
        \hline
        Id & Lasso & GB & Early Stopping & Lockout(L1) \\
        \hline
        
        \multirow{2}{*}{1} & $0.053\pm0.041$ & $0.072\pm0.057$ & $0.168\pm0.188$ & $0.054\pm0.042$ \\
        & \scalebox{0.9}{$(7.07\times10^{-1})$} & \scalebox{0.9}{$(3.28\times10^{-3})$} &  
        \scalebox{0.9}{$(5.48\times10^{-8})$} & \\
        \hline
        
        \multirow{2}{*}{2} & $0.068\pm0.070$ & $0.113\pm0.061$ & $0.177\pm0.099$ & $0.067\pm0.060$ \\
        & \scalebox{0.9}{$(8.79\times10^{-1})$} & \scalebox{0.9}{$(2.10\times10^{-10})$} &    \scalebox{0.9}{$(3.27\times10^{-20})$} & \\
        \hline
        
        \multirow{2}{*}{3} & $0.020\pm0.005$ & $0.019\pm0.005$ & $0.024\pm0.005$ & 
        $0.019\pm0.004$ \\
        & \scalebox{0.9}{$(3.66\times10^{-4})$} & \scalebox{0.9}{$(3.20\times10^{-1})$} &  
        \scalebox{0.9}{$(6.48\times10^{-22})$} & \\
        \hline
    \end{tabular}
    \caption{Mean error and standard deviation over 100 different partitions for each method. The paired t-test p-values between Lockout and the different methods are shown in parentheses in each case.}
    \label{tab:realdata1}
\end{table}

\begin{figure}[ht!]
    \centering
    \includegraphics[width=.7\textwidth]{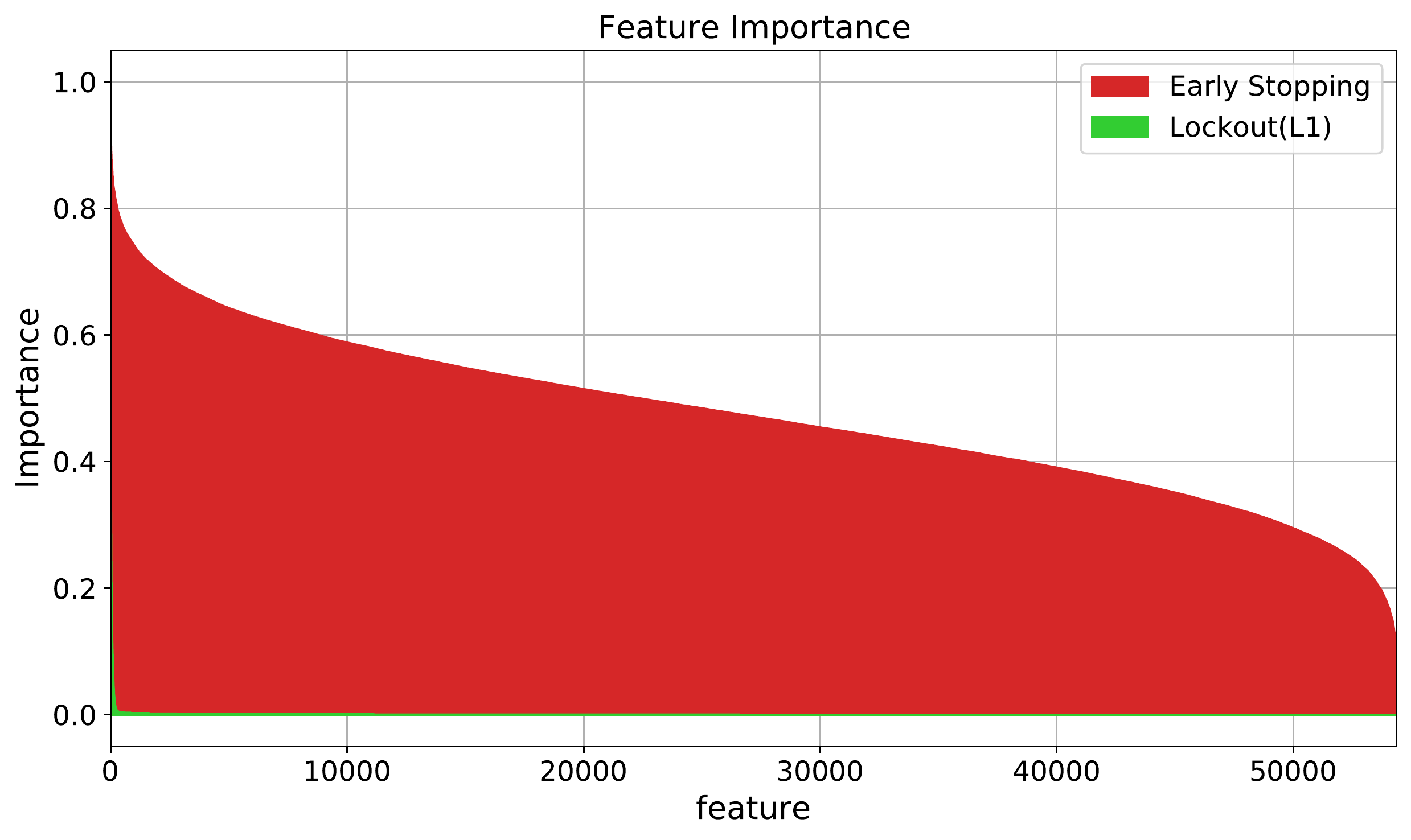}
    \caption{Feature importance, computed as $\sum\Omega_i/\mathrm{max}\{\Omega_i\}_1^{p}$ where $\Omega_i=\sum^{K}_{j=1}|w_{ij}|$ and $K$ is the number of nodes in the layer. Early stopping and Lockout(L1) models shown here correspond to one of the partitions on dataset 2.}
    \label{fig:feat_import_LO_2}
\end{figure}

 Table~\ref{tab:realdata1} shows the Mean error and standard deviation over 100 different partitions for each method. The paired t-test p-values between Lockout and the different methods are shown in parentheses in each case. Lockout(L1) is able to significantly improve the performance of the NN, making it comparable with state-of-the-art algorithms for tabular data such as Lasso or GB. Even more remarkable is the fact that this increase in performance came in connection with a dramatic reduction in the number of non-zero valued features as compared to the traditional early stopping solution. This can be appreciated in Figure~\ref{fig:feat_import_LO_2}, where feature importance graphs corresponding to early stopping and Lockout(L1) models for one of the partitions of dataset 2 are shown.

\subsubsection*{Experiment 2}

Three datsets from the UCI repository~\cite{Dua:2019} were analyzed, including two regression and one classification problem. They are summarized in Table~\ref{tab:exp2}. 
All experiments were performed following the same procedure and NN architecture described for Experiment~1. 
As seen in Table~\ref{tab:realdata2}, Lockout(L1) is able to improve the performance of the NN for the first two examples. In the last one (dataset 6), Lockout performs almost as well while reducing the number of non-zero features by roughly 81$\%$. In this case, we explored starting the algorithm at the early stopping rather than the unconstrained solution to investigate whether there exist solutions other than the validation error minimum that might be more desirable.

\renewcommand{\arraystretch}{1.4}
\begin{table}[ht!]
    \centering
    \begin{tabular}{|c|c|p{13.8cm}|}
        \hline
        Id & Size & \multicolumn{1}{|c|}{Description} \\
        \hline
        
        \multirow{2}{*}{4} & \multirow{2}{*}{2118$\times$124} & Communities and Crime Unnormalized dataset; the target variable being the total number of non-violent crimes per 100K population. \\
        \hline
        
        \multirow{2}{*}{5} & \multirow{2}{*}{372$\times$103} & Residential Building dataset; the target variable being the construction sale prices corresponding to single-family residential apartments in Tehran, Iran. \\
        \hline
        
        \multirow{2}{*}{6} & \multirow{2}{*}{200$\times$10000} & Arcene dataset; this is a two-class classification problem where the target is cancer versus normal patterns from mass-spectrometric data. \\
        \hline
        
    \end{tabular}
    \caption{Datasets from the UCI repository~\cite{Dua:2019} used in Experiment 2. Dataset size nomenclature is the same as in Table~\ref{tab:exp1}.}
    \label{tab:exp2}
\end{table}

\renewcommand{\arraystretch}{1.4}
\begin{table}[ht!]
    \centering
    \begin{tabular}{|c|c|c|c|c|c|}
        \hline
        Id & Lasso & GB & Early Stopping & Lockout(L1) \\
        \hline
        
        \multirow{2}{*}{4} & $0.682\pm0.049$ & $0.688\pm0.039$ & $0.683\pm0.043$ & $0.666\pm0.043$ \\
        & \scalebox{0.9}{$(1.48\times10^{-12})$} & \scalebox{0.9}{$(1.94\times10^{-21})$} &    \scalebox{0.9}{$(1.45\times10^{-18})$} & \\
        \hline
        
        \multirow{2}{*}{5} & $0.197\pm0.046$ & $0.212\pm0.047$ & $0.180\pm0.057$ & $0.164\pm0.042$ \\
        & \scalebox{0.9}{$(1.19\times10^{-13})$} & \scalebox{0.9}{$(9.82\times10^{-20})$} &    \scalebox{0.9}{$(9.69\times10^{-4})$} & \\
        \hline
        
        \multirow{2}{*}{6} & $0.289\pm0.067$ & $0.237\pm0.072$ & $0.239\pm0.082$ & $0.259\pm0.081$ \\
        & \scalebox{0.9}{$(3.84\times10^{-3})$} & \scalebox{0.9}{$(1.82\times10^{-2})$} &  
        \scalebox{0.9}{$(4.19\times10^{-2})$} & \\
        \hline
    \end{tabular}
    \caption{Mean error rate and standard deviation over 100 different partitions for each method. Nomenclature is the same as in Figure~\ref{tab:realdata1}.}
    \label{tab:realdata2}
\end{table}

\begin{figure}[ht!]
    \centering
    \includegraphics[width=.7\textwidth]{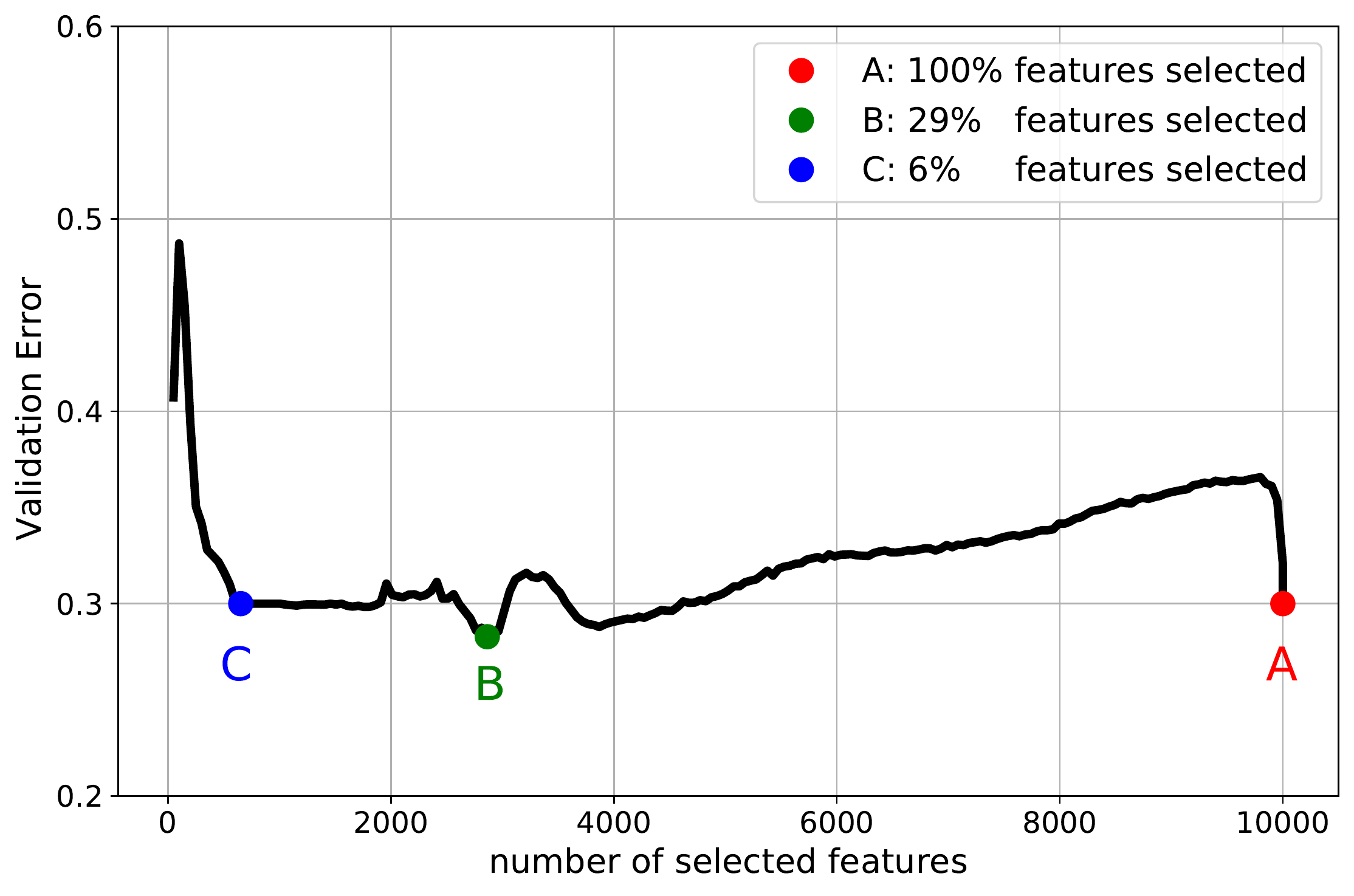}
    \caption{Validation Error vs. non-zero features in the model. This plot shows the validation error over the regularization path with Lockout on the dataset 6.}
    \label{fig:error_vs_features_sec5}
\end{figure}

The dependence of the validation error with number of selected features along this path is shown in Figure~\ref{fig:error_vs_features_sec5}. Point A represents the model chosen by early stopping at the validation minimum. As more regularization is applied, the NN performance initially degrades but then gradually improves both in terms of accuracy and feature selection, reaching a \emph{plateau} from around 4000 to 600 features. In this case, a solution (Point B) with better estimated performance than that of the early stopping solution was uncovered. However, selecting the model represented at point C might be preferred because it has similar estimated accuracy while involving only $6\%$ of the features. This phenomenon often occurs with smaller data sets that are more prone to noisy validation error estimates.

\subsection{Non-convex logarithmic regularization}

One of the intrinsic advantages of Lockout is that one can implement different regularization types by simply changing the constraint function and providing the new derivative, $p(\bm{w})$. This constraint need not be a convex function of the parameters. The only restriction is that it must be monotone, increasing with the absolute value of each parameter. Here, we illustrate this by employing a non-convex (logarithmic) constraint that provides more aggressive variable selection than the Lasso. This constraint is given by

$$
P_{\beta}(\bm{w}) = \sum_{j=1}^J \log((1-\beta) |w_j| + \beta); \quad 0<\beta<1,
$$
where $J$ is the number of parameters that will be included in the regularization. This regularization has the property that parameters with smaller absolute values are more aggressively penalized than larger ones as the strength of the constraint is increased. This can be seen by examining the derivative of $P$ respect to $|w|$: 

$$
p_{\beta}({w_j}) = \frac{1-\beta}{(1-\beta) |w_j| + \beta}
$$

\begin{figure}[ht!]
     \centering
     \begin{subfigure}[b]{0.49\textwidth}
         \centering
         \includegraphics[width=\columnwidth]{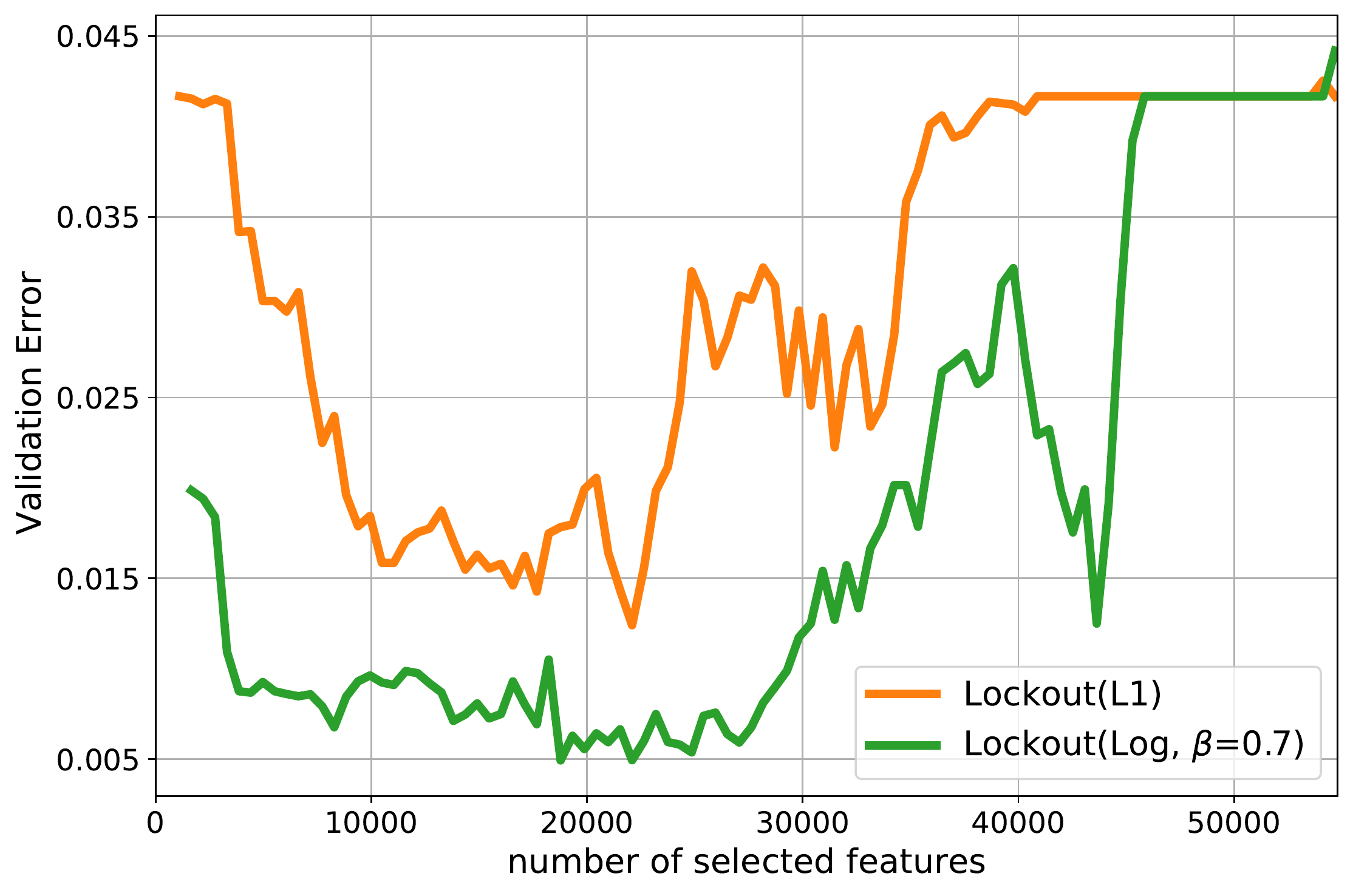}
         \caption{Dataset 1.}
         \label{fig:error_vs_density_b}
     \end{subfigure}
     \begin{subfigure}[b]{0.49\textwidth}
         \centering
         \includegraphics[width=\columnwidth]{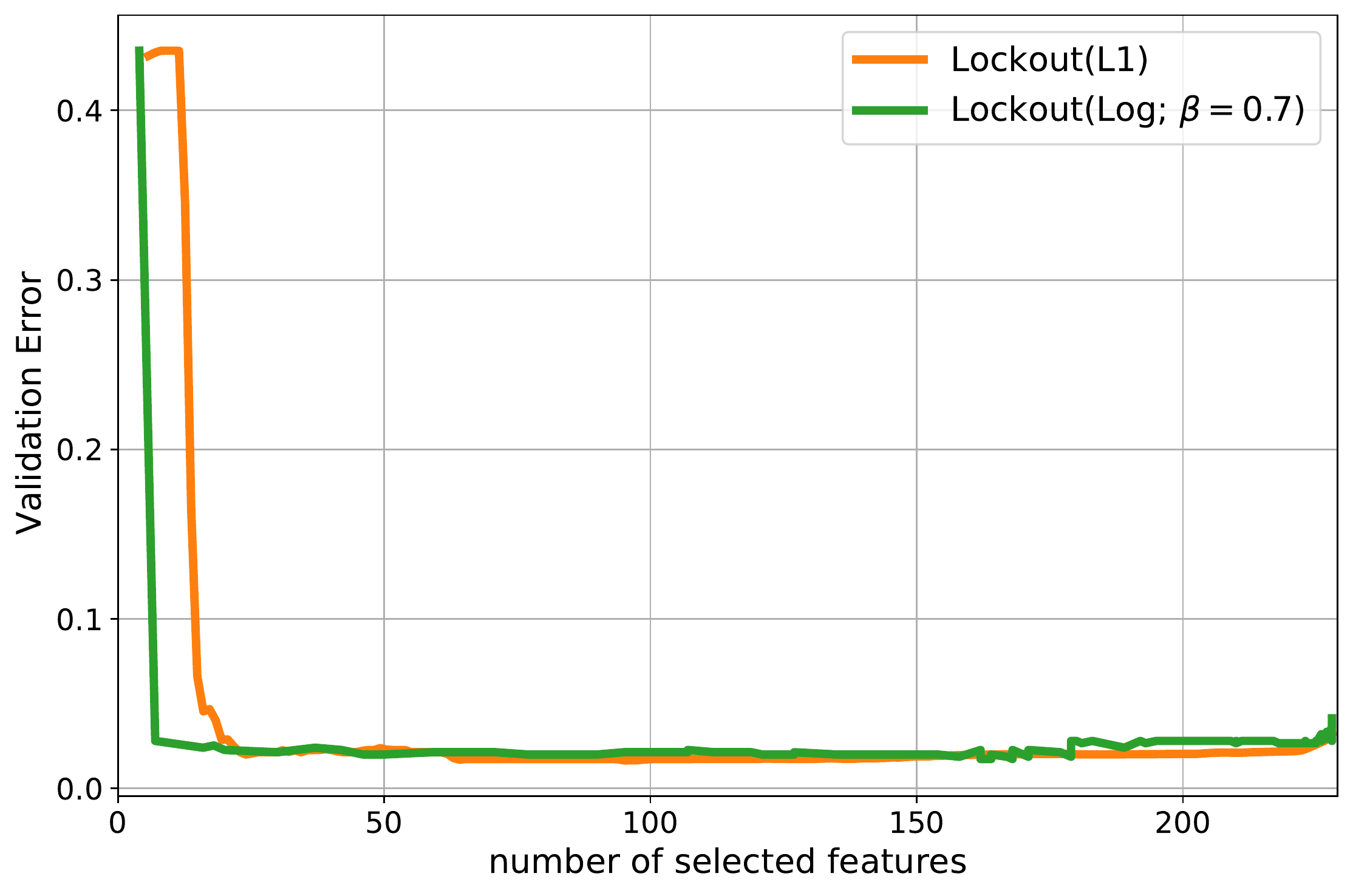}
         \caption{Dataset 3.}
         \label{fig:error_vs_density_a}
     \end{subfigure}
     \caption{Comparison of feature selection provided by Lockout with L1 or Logarithmic penalties. Results shown here are for one of the partitions of the corresponding dataset, using a 2-layer fully connected architecture with 10 nodes in the hidden layer followed by ReLU activation functions. As it can be seen, the Logarithmic penalty provided more aggressive feature selection than L1 which resulted in higher accuracy for the problems analyzed.}
     \label{fig:error_vs_density_log}
\end{figure}

One sees that the rate of reduction in absolute value of smaller valued parameters is increased relative to that of the larger ones, thereby emphasizing sparser solutions. This is illustrated in Figure \ref{fig:error_vs_density_log} by showing validation error versus the number of selected features in two different problems and contrasting it with the corresponding behavior of L1 regularization. Results shown in Figure~\ref{fig:error_vs_density_log} are for one of the partitions of the corresponding dataset, using a 2-layer fully connected architecture with 10 nodes in the hidden layer followed by ReLU activation functions. There were two nodes in the output layer. In both of these cases, more accurate and sparse solutions are obtained with the non-convex logarithmic regularization. 

\subsection{Common Representation of Cancer}

In this section, we explore a practical application of Lockout to cancer research. Integration of Gene Expression datasets (meta study) has proven extremely valuable as these datasets tend to have some observations with many genes, as described below. Meta studies to date have focused on finding a gene, single set of genes or pathways that are over expressed in a single phenotype (e.g. breast cancer) \cite{nguyen2019identifying}. Specifically, the studies of over expressed pathways  have been shown to be more robust than the analysis of single genes because the individual genes seldom act alone \cite{nguyen2019identifying}. Today, however, these pathways are predefined by the investigators in the majority of cases \cite{subramanian2005gene}. 

\begin{figure}
     \centering
     \begin{subfigure}[a]{.7\textwidth}
         \centering
         \caption{Multi Target NN. Lockout is applied to the first layer to provide L1 regularization. The Hidden Layer vector is learned is common to all the problems }
         \includegraphics[width=\columnwidth]{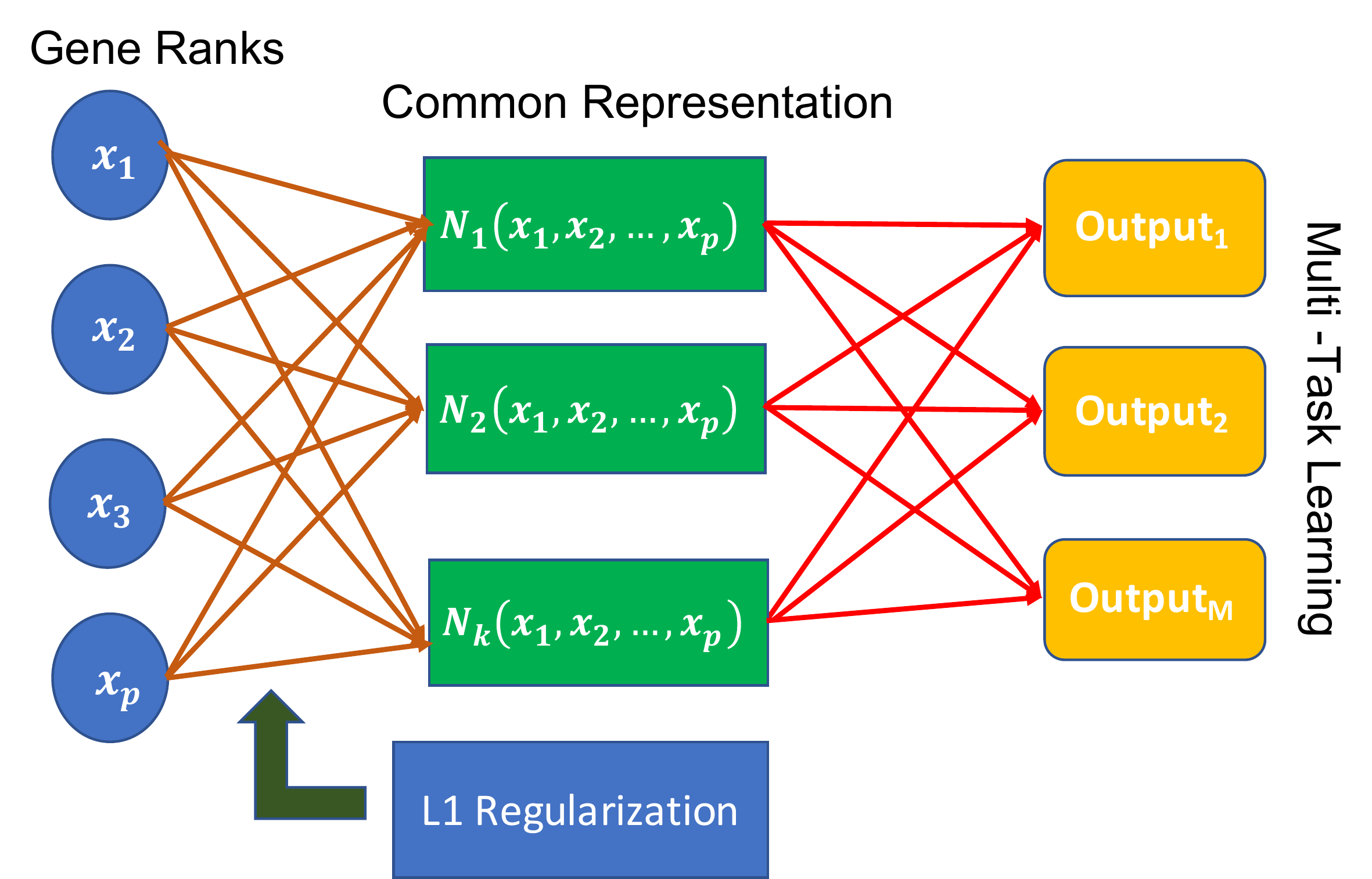}
         \label{fig:common_representation_lockout_architecture}
     \end{subfigure}
     \begin{subfigure}[b]{.7\textwidth}
         \centering
         ~\\
         \caption{Lockout applied to 585 different problems to provide regularization in the first layer. Point A represents the early stopping solution from where Lockout starts an point B is the minimum found.}
         \includegraphics[width=\columnwidth]{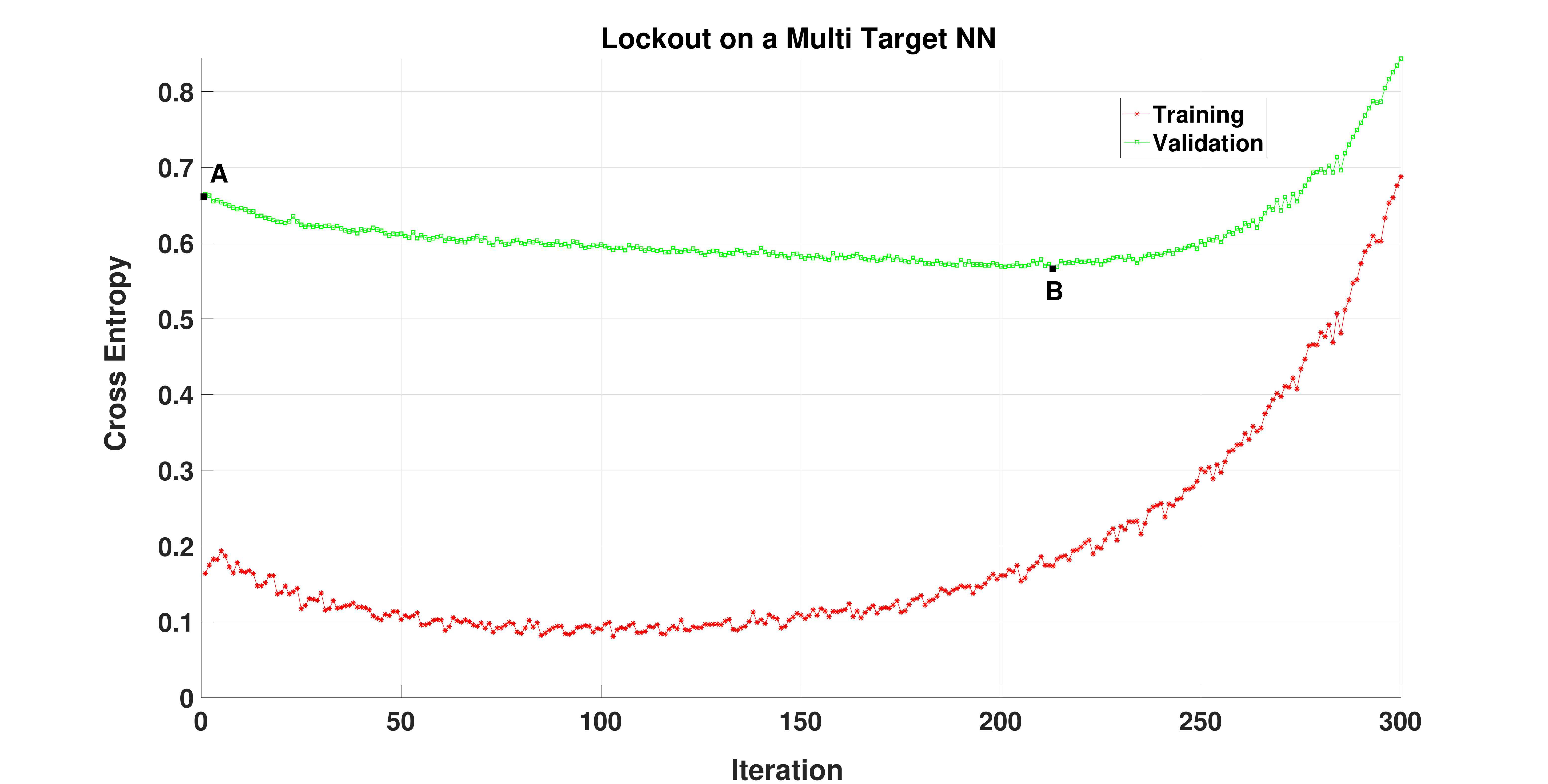}
         \label{fig:lockout_common_representation}
     \end{subfigure}
\end{figure}

We were interested in designing an algorithm capable of learning which genes to chose and how to combine them (synthetic pathways) in order to explain a wide range of cancer  problems and phenotypes. We took advantage of combining representational learning with Lockout to design a multi-output Neural Network with a common hidden layer, as shown in Figure \ref{fig:common_representation_lockout_architecture}. The function of the hidden layer is to transform the original inputs into a space where many problems are easily solvable (the common representation). We expect that by borrowing strength from multiple problems, a better representation is obtained that can improve the accuracy of each individual problem.

In order to implement our architecture, we require datasets with the same sets of genes. We chose datasets belonging to the L1000 platform in the GEO BioBank repository \cite{subramanian2017next} that contain expressions of at least 100 of the 978 genes. We chose the genes in the L1000 Platform because Subramanian et al \cite{subramanian2017next} defined them in a data driven approach. They maximized the information gained about the whole transcriptome while minimizing costs \cite{subramanian2017next}. They showed that this set of landmark genes can explain up to 81\% of all genes that are not measured \cite{subramanian2017next}. 

As outcomes for each of these expression profiles, we need the phenotype columns that can be used for prediction (e.g. cancer vs no cancer, metastasis vs no metastasis). Since the GEO Biobank was not created with the purpose of being analyzed with Machine Learning methods, we created a pipeline to extract automatic phenotypes from these datasets. We were able to effectively do so by combining NLP analysis, data preprocessing pipelines and manual curation (assisted by a trained group of data science master students). We assembled 553 different datasets across 23 different cancer types (The 5 Major groups are Breast = 208, Colorectal = 89, Lung = 54, Prostate = 52, Ovarian Cancer = 36). The phenotypes range from different type of cells, aggressiveness of the tumors, cancer types, metastasis status, etc.  All of these datasets have at least 5 observations per category and can be downloaded at \url{https://lifelong-ml.github.io/madna.github.io/}. Finally, each selected study uses only one platform to facilitate internal normalization and batch correction using ranks before the multitask learning analysis. 

In total, we used 39516 observations to learn relationships among 978 genes for 23 major classes of cancer. The data was divided in 70\% for training, 10\% for validation and 20\% for testing. Only datasets with at least 20 observations in the test set were used for testing.  There were 500 nodes in the hidden layer using Relu as activation function and the learning rate was cross validated in every case between the values $\left[ 0.0001, 0.001, 0.01, 0.1 \right]$. The Adam optimizer was used to find the minimum in the validation set in the forward step. Lockout with L1 regularization was applied to the first layer, starting at the minimum of the validation solution. The sum of the absolute values of the coefficients for each of the output layers was not allowed to change. 

As it can be seen in Figure \ref{fig:lockout_common_representation}, the use of Lockout improved the relative performance of the  forward network (initial point) by approximately $14\%$. Additionally, to test whether the combined representation learned had meaning, we performed an experiment using all datasets with more than 100 observations (67 datasets). From the combined network we 1) extracted the coefficients of the hidden layer and 2) for each problem we extracted the 67 set of coefficients of the output layers. We then connected them, creating 67 distinct NNs. Each of this NNs was used as starting point to find the minimum at validation for their specific problem, and the performance on the test set was recorded. Additionally, we also optimized NNs for each problem with the same architecture, but the parameters were started randomly. The NNs starting from the common representation significantly improved performance in 74\% of the datasets. The median error improvement across all dataset was of 18\%.  The null hypothesis that starting with the representation obtained by Lockout offers the same performance as initializing the parameters randomly was rejected with a $p = 7\times10^{-11}$ using a paired t-test. 

Equally relevant to the improvement in performance is the use of Lockout to identify the most important genes that help explain the many cancer problems. Table \ref{tab:common_representation} shows the 5 most important genes selected by Lockout. Interestingly, all of these genes are related to hallmarks of cancer: cell proliferation, survival mechanism and energy production. An interesting analysis, although beyond the scope of this article, would be to check the genes with similar strength at the different nodes to discover unknown connections and interactions. Equally interesting would be to detect which specific genes are related to the different cancers. 

\renewcommand{\arraystretch}{1.4}
\begin{table}[ht!]
    \centering
    \begin{tabular}{|c|p{8.3cm}|p{5.5cm}|}
   
        \hline
         \textbf{Gene Name} & \textbf{Function} & \textbf{Known Diseases Associated} \\
        \hline
        
        CCNE2 & Essential for the control of the cell cycle at the late G1 and early S phase. & T-Cell Lymphoma of Childhood and Retinoblastoma. \\
        \hline
        
        SFN & This gene encodes a cell cycle checkpoint protein. The encoded protein binds to translation and initiation factors and functions as a regulator of mitotic translation.  &  Benign Breast Adenomyoepithelioma and Colloid Carcinoma Of The Pancreas. \\
        \hline
        
        PGAM1 & It is a fundamental gene involved in the glycolytic pathway which creates the energy sources adenosine triphosphate (ATP) and nicotinamide adenine dinucleotide (NADH). & Menkes Disease and Myoglobinuria. \\
        \hline
        
         CIRBP & Cold-inducible mRNA binding protein that plays a protective role in the genotoxic stress response by stabilizing transcripts of genes involved in cell survival.  & Cryptorchidism, Unilateral Or Bilateral and Acth-Secreting Pituitary Adenoma. \\
         \hline
        
        GPR56 & This gene encodes a member of the G protein-coupled receptor family and regulates brain cortical patterning. The encoded protein binds specifically to transglutaminase 2, a component of tissue and tumor stroma implicated as an inhibitor of tumor progression. &  Polymicrogyria, Bilateral Frontoparietal and Polymicrogyria, Bilateral Perisylvian, Autosomal Recessive. \\
        \hline
        
    \end{tabular}
     \caption{ 5 Top most important genes and their functions in order to explain all cancer phenotypes obtained. Three of genes are related to cell and tumor proliferation, one to cell survival and the other one to energy production. All these processes are hallmarks of cancer.}
    \label{tab:common_representation}
\end{table}

\section{Discussion}

In this paper we have introduced a general regularization method (Lockout) for linear and non-linear models, including Neural Networks, with sparsity inducing constraint functions. The crux of the algorithm is to repeatedly linearize the loss function and the constraint, and then solve the resulting linear programming problem.

This approach offers several advantages. There is no Lagrangian in the loss function. The gradients are not changed, and the procedure does not stop learning. It can be applied with any differentiable model function and loss, and at any (or all) layers of any architecture. It can be driven by any constraint that is a differentiable monotonically increasing  (convex or non-convex) function of the absolute value of each parameter. It can be applied with any existing optimization method (e.g. Adam, Stochastic Gradient Descent, etc.) and is easily implemented with a few lines of code. Compared to regular back propagation it only increases the computational complexity as $o(\bm{M}\log\bm{M})$ so it can be effectively applied in practice. Finally, it navigates the entire path with only a single optimization run. This eliminates the need to employ repeated cross-validation runs in order to estimate the optimal value of the regularization parameter. 

In practice, Lockout also offers some distinct advantages. As illustrated here, it usually outperforms the early stopping NN solution while at the same time providing significant feature selection and complexity reduction. This is the case even in tasks for which NNs were often considered to be inferior choices (e.g. for problems for which the number of features was substantially larger than the number of observations). It also caused NNs to be more competitive in the analysis of tabular data, sometimes improving performance over state of the art algorithms like Lasso or Gradient Boosting. As such, automatic feature engineering and representational learning can be incorporated into problems for which they were not considered to be amenable, like the common representation of cancer. Moreover, the simplicity of incorporating different constraint functions allows convenient experimentation with a variety of regularization alternatives. Finally, when initialized at the usual early stopping solution, it is guaranteed (in expectation) to perform at least as well while usually simultaneously improving accuracy and sparsity. This opens the door to the use of Lockout on existing already deployed NNs to potentially improve accuracy and interpretation.  

For future work, we would like to study the Group Lasso, or other constraints with similar properties, as applied to NNs. There are several natural ways to group parameters in NNs each enforcing different interesting behavior. For instance, one could arrange all the input parameters at each of the different nodes to be together in the same group. This will enforce architecture selection by enabling and disabling the various nodes as the regularization path is traversed. Similarly, grouping together all first layer parameters connected to the same input variable might result in stronger regularization, since each variable must enter all the nodes at the same point in the regularization path. Finally, it might be  interesting to study the use of Lockout as a possible competitor to attention mechanisms \cite{vaswani2017attention} which presently offer state-of-the-art results on imaging and text processing tasks. 

PyTorch code and documentation for Lockout and all the experiments performed is provided at \url{https://github.com/warbelo/Lockout.git}.

\pagebreak
\printbibliography

\pagebreak

\section{Appendix A. Definitions}

\setcounter{equation}{0}

Consider the optimization problem:
\begin{equation}
\begin{array}{rrclcl}
\displaystyle \hat{\bm{w}} = \argmin_{\bm{w}}  \mathbb{E}_{\bm{x},y} \left( L\left[y, f\left(\bm{x};\bm{w}) \right)\right]\right)
& \textrm{s.t.} & P(\bm{w})\leq t_0\\
\end{array}
\label{eqn1appendix}
\end{equation}
where $L$ is a loss function, $f$ is a non-linear function (e.g. NN), and  $P(\bm{w})$ is a general constraint function that monotonically increases with $|\bm{w}|$ so that 
\begin{equation*}
    \quad\frac{\partial P(\bm{w})}{\partial |{w}|} \geq 0.
\end{equation*}
For instance, $P(\bm{w}) = \sum |w|$ for Lasso and $P(\bm{w}) = \sum w^2$ for Ridge regression.

Let us define:
\begin{align*}
    g_{j} &= -\frac{\partial \mathbb{E}_{\bm{x},y} \left( L\left[y, f\left(\bm{x},\bm{w}) \right)\right]\right)}{\partial w_{j}}, \\
    p_{j} &= \frac{\partial P(\bm{w})}{\partial|w_{j}|}, \\
    \gamma_{j} &= \frac{|g_{j}|}{p_{j}}
\end{align*}
where $j$ is a generic parameter indexing the weights. Note that according to our definitions $\gamma_{j} \geq 0$. Finally, we \textbf{bold} letters for vectors and regular letters for its component.

\subsection{Appendix B. A first order Generalized Regularization step}

Let us assume that we start with parameters equal to $\bm{w}^0$. For linear models, usually the parameters are initialized at a solution of Eq.~\ref{eqn1appendix} by setting all of them equal to zero. This initialization is not possible for NNs as all the gradients would be zero valued. Additionally, they can not be initialized too close to zero either because the gradient would be too small, and the algorithm would not learn. For NNs they can be initialized to the values of the unconstrained solution as this solution is on the regularization path.

Now, let us focus on finding an approximate solution to Eq.~\ref{eqn1appendix} for a given $t_0$. Suppose the values of the parameters $\bm{w}$ at iteration $i$, $\bm{w}^i$, are known, and we want to update them with $\Delta \bm{w}$. First, we linearize $L$ and $P$ using a first-order Taylor expansion around $\bm{w}^i$ and transform the initial optimization problem into\footnote{For reasons that will be obvious later on, the following chain rule property $\frac{\partial P(\bm{w})}{\partial w_{j}} = \frac{\partial P(\bm{w})}{\partial|w_{j}|} \frac{d|w_{j}|}{w_{j}}$ has been used when linearizing the constraint.} 
\begin{align*} 
    \max_{\Delta \bm{w}} \sum\limits_{j} g_{j}\Delta w_{j} \quad \textrm{s.t.} \sum\limits_{w_{j}=0} p_{j} |\Delta w_{j}| + \sum\limits_{w_{j} \neq 0} p_{j} \mathrm{sign}(w_{j}^i) \Delta w_{j}  &\;\leq\; t_0 - P(\bm{w}^i)\\
    |\Delta w_{j}|  &\;\leq\;  s_{j},
\end{align*}
where $s_{j}$, that define the trust region, is the absolute value of the step size for each parameter for which the first-order Taylor expansion for both the objective and constraint functions are valid. Therefore, if we solve the linear programming problem above by finding the optimal $\Delta \bm{w}$ we can have a solution of Eq.~\ref{eqn1appendix} in the vicinity where the first-order Taylor approximation is valid, $\bm{w}^{i+1} = \bm{w}^i + \Delta \bm{w}$ . By applying these steps iteratively, we can then solve the original problem. 

Please note that those parameters for which $p_j = 0$ are removed from the constraint because this term shows up in both summands. Therefore, they can be directly updated to their unconstrained value, $\Delta w_{j}=\mathrm{sign}(g_{j}) s_{j}$. This behavior explains why methods with $p_j = 0$ when $w=0$ (e.g. L2) do not perform feature selection. As soon as the parameter is 0 it will be updated in the next iteration because it does not have penalization. This is not the case of L1 or Group Lasso regularization. Removing the parameters for which $p_j = 0$ the new optimization problem can be written as:

\begin{align*} 
    \max_{\Delta \bm{w}} \sum\limits_{j} g_{j}\Delta w_{j} \quad \textrm{s.t.} \sum\limits_{w_{j}=0, p_j \neq 0} p_{j} |\Delta w_{j}| + \sum\limits_{w_{j} \neq 0, p_j \neq 0} p_{j} \mathrm{sign}(w_{j}^i) \Delta w_{j}  &\;\leq\; t_0 - P(\bm{w}^i)\\
    |\Delta w_{j}|  &\;\leq\;  s_{j},
\end{align*}

This formulation results in a linear programming problem with an exact solution that depends on the coefficient values. Let us introduce a change of variables given by $\Delta w_{j} = \frac{\mathrm{sign}(g_{j})}{p_j} m_{j}$ which results in 

\begin{align*} 
    \max_{m_{j}} \sum_{j} \gamma_j m_{j} \qquad\textrm{s.t.} \sum\limits_{w_{j}=0; p_j \neq 0} |m_{j}|\quad + \sum\limits_{w_{j} \neq 0; p_j \neq 0} \mathrm{sign}(w_{j}^i) \mathrm{sign}(g_{j}) m_{j}  &\;\leq\; t_0 - P(\bm{w}^i)\\
    | m_{j}|  &\;\leq\;  p_{j}s_{j}.
\end{align*}
From now on we will omit the condition $p_j \neq 0$ from the summations. It is implicit that the analysis that follows does not apply to them and these parameters should be updated to the unconstrained value. 

Let us assume for reasons that will become clear below that all the coefficients in the summations are sorted in descending order according to the values of $\gamma_{j}$ \footnote{This step is the one that defines the computational complexity of our algorithm as $o(\bm{M}\log\bm{M})$, where $M$ is the number of parameters being regularized.}. Let $SS$ be the set of parameters where $ \mathrm{sign}(w_{j}^i)=\mathrm{sign}(g_{j})$ and $w_{j} \neq 0$. Let $DS$ be  those for which $ \mathrm{sign}(w_{j}^i)\neq\mathrm{sign}(g_{j})$ and $w_{j} \neq 0$. We can then write the above optimization problem as
\begin{align*} 
    \max_{m_{j}} \sum_{j} \gamma_j m_{j} \qquad\textrm{s.t.}\quad \sum\limits_{w_{j}=0} |m_{j}| \;+\; \sum\limits_{SS} m_{j}  -\sum\limits_{DS} m_{j} &\;\leq t_0\; - P(\bm{w}^i)\\
    | m_{j}|  &\;\leq\;  p_{j}s_{j}.
\end{align*}
Note that for those parameters belonging to $DS$, $m_{j} = p_{j}s_{j}$ (both  $\gamma_j$ and $m_{j}$ are positive numbers)  since increasing them to their maximum allowed value increases the objective function the most  while increasing the slack for other parameters. This condition is equivalent to letting all the parameters that would like to decrease their absolute values while at the same time increase the objective function to do it. After setting these parameters to their respective values, the optimization problem for $m_j$ where $j \in DS^c$, can be expressed as
\begin{align*} 
    \max_{m_{j}} \sum\limits_{DS^c} \gamma_j m_{j} \qquad\textrm{s.t.}\quad \sum\limits_{w_{j}=0} |m_{j}| + \sum\limits_{SS} m_{j}   \leq t_0 - P(\bm{w}^i)  + t_{DS} \\
    | m_{j}| &\leq p_{j} s_{j},
\end{align*}
where $DS^c$ is complementary with $DS$ and $t_{DS} = \sum\limits_{DS} p_{j} s_{j}$.

In order to maximize the objective function above, we would like to start filling in the parameters in $DS^c$ that have the largest $\gamma_j$ with the values $m_j = p_{j}s_{j}$. This in turn will increase the left side of the first constraint and might have to be compensated by setting others to $m_j = - p_{j}s_{j}$ (those with the smallest value of $\gamma_j$ whose $w_{j} \neq 0$). The extent of this compensation  will depend on the term $t_0 - P(\bm{w}^i)  +  t_{DS}$. 

Let us define
\begin{equation*} 
    \Delta_{J^*} = \sum_{\substack{j=J^{*} + 1 \\ w_{j} \neq 0 }}^{J} p_{j}s_{j} +  t_0 - P(\bm{w}^i)  +  t_{DS}  - \sum_{\substack{j=1}}^{J^{*}-1} p_{j} s_{j} \;\;
    \text{for} \;\; J^* \in \{1 \dots J\}
\end{equation*}
where $J$ is the total number of parameters in $DS^c$, again sorted.
It can be shown by checking the KKT conditions that
\begin{equation} 
    m_{J^*} = 1(w_{J^*}  \neq 0)^{1-\mathrm{sign}(\Delta_{J^*})} \mathrm{sign}(\Delta_{J^*}) \min(p_{J^*}s_{J^*} , |\Delta_{J^*}|)  \;\;
    \text{for} \;\; J^* \in \{1 \dots J\}
\label{eqn3appendix}
\end{equation}
is the solution to the problem above. Let us examine Eq.~\ref{eqn3appendix} for better understanding. First, note that that $\Delta_{J^*}$ determines the size of the step of the parameter $J^*$ so that the constraint 1 is met. If $\Delta J^* > 0$ then, $m_{J^*} = \min(p_{J^*}s_{J^*} , |\Delta_{J^*}|)$ which is the maximum step size that still meets both constraints. If $\Delta J^* \leq 0$ and $w_{J}^* \neq 0$ then, $ m_{J^
*} = - \min(p_{J^*}s_{J^*} , |\Delta_{J^*}|) $ which means that the parameter will take the maximum step necessary in the negative direction to meet both constraints and compensate other parameters whose step is positive. Finally, if  $\Delta J^* \leq 0$ and $w_{J}^* = 0$ then $m_{J^
*} = 0$ because taking any step, positive or negative, will increase the constraint. This is the behaviour that results in feature selection if $p(w) \neq 0$ when $w = 0$. Changing to the original variables, the solution can then be written as\\
\textbf{Solution:}
\begin{align*}
  \mathbf{if}\quad & p_{J} = 0\quad\mathbf{or}\quad(\mathrm{sign}(w_{J}) \neq \mathrm{sign}(g_{J})\quad\mathrm{and}\quad w_{J} \neq 0): \\
  \\
  &\Delta w_{J} =  \mathrm{sign}(g_{J}) s_{J} \\
  \\
  \mathbf{else} & \\
  &\Delta w_{J} = 1(w_{J} \neq 0)^{1-\mathrm{sign}(\Delta_{J})} \mathrm{sign}(g_{J}) \mathrm{sign}(\Delta_{J}) \min(s_{J} , \frac{|\Delta_{J}|}{p_{J} + \epsilon})
\end{align*}
Therefore, up to a first order approximation of the loss and constraint functions, the solution above is a solution to problem  \ref{eqn1}. Applying this solution iteratively, we can find the solution to the original problem.

\subsection{Appendix C. Traversing the Regularization Path}

Let $ t_0 = P(\bm{w})  + \Delta t$. Note that depending on the value of $\Delta t$ the constraints
\begin{equation*}
    \sum\limits_{w_{j}=0} |m_{j}| + \sum\limits_{SS} m_{j} \;\leq\; \Delta t  +  t_{SD} \quad\qquad\mathrm{and}\quad\qquad |m_{j}|\;\leq\;p_{j} s_{j}
\end{equation*}
might be impossible to be simultaneously met. In this case we would set all $m_{j} = -  p_{j} s_{j}$ and the only parameters that would move in the direction of decreasing the loss functions are those that belong to $DS$. This will continue until both constraints can be satisfied, in which case the algorithm will start trading an increase of the absolute value of the most important parameters with the reduction of those that are less important until the solution is reached. If we assume that $\bm{w}$ is the solution when $t_0 = P(\bm{w})$ and $\Delta t$ is taken to be small enough so that both constraints can be met within the boundaries of the first order Taylor approximations, then the parameters $\bm{w} = \bm{w} + \Delta \bm{w}$ will be the solution to the problem when $t_0 = P(\bm{w}) + \Delta t$. This allows the design of a simple algorithm to traverse the path. Start with the unconstrained solution $\bm{w} = \bm{w}^{uc}$. Then set $t_0 = P(\bm{w}) - \Delta t $ to traverse the reverse path where $\Delta t >0 $ and $|\Delta t| < \epsilon$ being $\epsilon$ a small enough number. Note that if $\Delta t$ is not small then many steps may be needed to reach the solution, but the same algorithm can be used. 

\subsection{Appendix D. Choosing $s_j$}

As mentioned above, $s_j$ is defined to be a step size within which both Taylor approximations are still valid. If $s_j$ is taken from an optimizer, such as stochastic gradient descent or Adam, we can say that their value decreases the loss function and as such we will assume that they satisfy the linear approximation. We will also assume that these steps satisfy the Taylor approximation of the constraint, with a notable exception. If the step size is such that the parameters $w_j$ will change sign, because $p_j$ is a non-decreasing positive function of, $|w|$ then the first order Taylor approximation would not be valid for the constraint. In that case, those parameters are set to zero. They remain zero valued until they are selected to be among those that allowed to change, thereby giving rise to the name "Lockout". 

\subsection{Appendix E. Pseudocode}

Here we provide the pseudocode for Lockout. Please note that the same can be applied independently to any layer or the whole network together.

\begin{algorithm}[t]
  \footnotesize
\LinesNotNumbered
\DontPrintSemicolon 
\KwIn{
        \begin{itemize}
            \item $\{x_n, y_n\}_1^N$: training data.
            \item $I$: number of iterations, $\epsilon$: a small constant
            \item $NN$: network architecture
            \item $\bm{w}^{uc}$: weight from an unconstrained solution 
            \item $\nu$: learning rate
        \end{itemize}
}
\KwOut{$NN$ }

~\\

- calculate $t_0 = P(\bm{w}_1^{uc})$ 

~\\

\For{i=1:I}{

~ \\
 
- Calculate  $\bm{g} = - \frac{d L}{d \bm{w} } \biggr\vert_{\bm{w} = \bm{w}^i}$ ,  $\bm{p} = \frac{d P}{d|\bm{w}| } \biggr\vert_{\bm{w} = \bm{w}^i}$, and $\Delta \bm{w} = \bm{g}$.

 ~\\
 
 \For{ Parameters in the first layer}{
~\\
~\\
- Define $DS$ as the set of parameters for which $p_{J} = 0$  or  $ \mathrm{sign}(w_{j}^i)\neq\mathrm{sign}(g_{j})$ and $w_{j}^i \neq 0$\\
~\\
\textbf{if} $J \in DS$
~\\
~\\
 $\quad\Delta w_{J} =   g_{J}$ \\ 
 ~\\
\textbf{else}
 ~\\
 ~\\
 
- let $t_{DS} = \sum_{j \in DS} p_j |g_j|$\\
 ~\\
- Calculate $\bm{\gamma} = \frac{|\bm{g}|}{\bm{p} + \epsilon}$ and sort them in descending order. 
 ~\\
~\\
- For $j \in DS^c$,  re-order $\Delta \bm{w}, \bm{g}$, and $\bm{p}$ to match the order in $\bm{\gamma}$:
~\\
~\\\quad - Calculate $\Delta_{J} =  \nu \sum\limits_{ \substack{j={J + 1} \\ w_{j} \neq 0}}^{J} p_{j}|g_{j}| +  t_0 - P(\bm{w}^i)  +  \nu \cdot t_{DS}  - \nu \sum\limits_{\substack{j=1}}^{J^{*}-1} p_{j}  |g_{j}|$ 
~\\
~\\

\quad - $\Delta w_{J} = 1(w_{J}^i \neq 0)^{1-\mathrm{sign}(\Delta_{J})} \mathrm{sign}(g_{J})\mathrm{sign}(\Delta_{J}) \mathrm{min}( |g_{J}| , \frac{|\Delta_{J}|}{ \nu (p_{J}+ \epsilon)})$  \\  
~\\

\textbf{end} \\
~\\
}
~\\

- Update the parameters $\bm{w}^{i+1} = \bm{w}^{i} - \nu \Delta \bm{w}$ \\

~\\

- \textbf{if} $\textrm{sign}(w_j^{i+1})\times\textrm{sign}(w_j^{i}) = -1:  w_j^{i+1} =0$ \\
  
~\\
~\\
}
\Return{$NN(\bm{w}^{I})$}
\caption{{Lockout. \textbf{(LO)}}}
\label{algorithm1}
\end{algorithm}

\end{document}